\documentclass{article}

\usepackage{arxiv}

\usepackage[utf8]{inputenc} 
\usepackage[T1]{fontenc}    
\usepackage{hyperref}       
\usepackage{url}            
\usepackage{booktabs}       
\usepackage{amsfonts}       
\usepackage{nicefrac}       
\usepackage{microtype}      
\usepackage{lipsum}
\usepackage{graphicx}       %

\title{Single Unit Status in Deep Convolutional Neural Network Codes for Face Identification:\\Sparseness Redefined}

\author{
  Connor J. Parde \\
  School of Behavioral and Brain Sciences\\
  The University of Texas at Dallas\\
  Richardson, TX 75080 \\
  \texttt{Connor.Parde@utdallas.edu} \\
   \And
 Y. Ivette Col\'on \\
  School of Behavioral and Brain Sciences\\
  The University of Texas at Dallas\\
  Richardson, TX 75080 \\
  \texttt{icolon@utdallas.edu} \\
     \And
 Matthew Q. Hill \\
  School of Behavioral and Brain Sciences\\
  The University of Texas at Dallas\\
  Richardson, TX 75080 \\
  \texttt{matthew.hill@utdallas.edu} \\
     \And
 Carlos D. Castillo \\
  UMIACS\\
  University of Maryland\\
  College Park, MD 20740 \\
  \texttt{carlosc@umd.edu} \\
     \And
 Prithviraj Dhar \\
  UMIACS\\
  University of Maryland\\
  College Park, MD 20740 \\
  \texttt{pdhar@cs.umd.edu} \\
     \And
 Alice J. O'Toole \\
  School of Behavioral and Brain Sciences\\
  The University of Texas at Dallas\\
  Richardson, TX 75080 \\
  \texttt{otoole@utdallas.edu}
}

\begin{document}
\maketitle

\begin{abstract}
Deep convolutional neural networks (DCNNs) trained for face identification develop representations that generalize over variable images, while retaining subject (e.g., gender) and image (e.g., viewpoint) information. Identity, gender, and viewpoint codes were studied at the ``neural unit'' and ensemble levels of a face-identification network. At the unit level, identification, gender classification, and viewpoint estimation were measured by deleting units to create variably-sized, randomly-sampled subspaces at the top network layer. Identification of 3,531 identities remained high (area under the ROC approximately 1.0) as dimensionality decreased from 512 units to 16 (0.95), 4 (0.80), and 2 (0.72) units. Individual identities separated statistically on every top-layer unit. Cross-unit responses were minimally correlated, indicating that units code non-redundant identity cues. This ``distributed'' code requires only a sparse, random sample of units to identify faces accurately. Gender classification declined gradually and viewpoint estimation fell steeply as dimensionality decreased. Individual units were weakly predictive of gender and viewpoint, but ensembles proved effective predictors. Therefore, distributed and sparse codes co-exist in the network units to represent different face attributes. At the ensemble level, principal component analysis of face representations showed that identity, gender, and viewpoint information separated into high-dimensional subspaces, ordered by explained variance. Identity, gender, and viewpoint information contributed to all individual unit responses, undercutting a neural tuning analogy for face attributes. Interpretation of neural-like codes from DCNNs, and by analogy, high-level visual codes, cannot be inferred from single unit responses. Instead, ``meaning'' is encoded by directions in the high-dimensional space.
\end{abstract}


\section{Introduction}
The concept of a {\it feature} is at the core of psychological and neural theories of visual perception. The link between perceptual features and neurons has been a fundamental axiom of visual neuroscience since Letvin et al. (1959) first described the receptive fields of ganglion cells as ``bug perceivers'' (pp. 258 \cite{lettvin1959frog}). At low levels of visual processing, features can be defined in precise terms and located in a retinal image. They can also be interpreted semantically (e.g., vertical line, retinal location $x$). These codes are sparse, because they rely on the responses of a small number of specific neurons \cite{olshausen1996emergence,olshausen1997sparse}. At higher levels of visual processing, where retinotopy gives way to categorical codes, the connection between receptive fields and  features is unclear. Although ``face-selective'' may be an accurate description of a neuron's receptive field, it provides no information about the features used to encode a face.

A fundamental difference between retinotopic representations in early visual areas and the categorical representations that emerge in ventral temporal cortex is that the latter generalize across image variation (e.g., viewpoint). Historically, face recognition algorithms relied on feature detection strategies analogous to those used in low-level vision (e.g., \cite{riesenhuber1999hierarchical,riesenhuber2000models}). These algorithms operated accurately only on controlled face images with limited variation in viewing conditions. Since 2014, face-identification algorithms based on deep convolutional neural networks (DCNNs) have largely overcome the limit of recognizing faces using image-based similarity \cite{sun2014deep,taigman2014deepface,sankaranarayanan2016triplet,schroff2015facenet,chen2015end,ranjan2017all}. Similar to face codes in high-level visual cortex, DCNN codes generalize  over substantial image variation. Indeed, the response properties of neurons in inferior temporal cortex can be simulated using appropriately weighted combinations of output units from a DCNN trained for object recognition \cite{yamins14neuralpredictions}. 

The parallels between primate vision and deep-learning networks are by design \cite{o2018face, Sejnowski201907373}. DCNNs employ computational strategies similar to those used in the primate visual system \cite{fukushima1988neocognitron,krizhevsky2012imagenet} and are trained extensively with real-world images. For face identification, effective training sets consist of variable images of a large number of identities. DCNNs process images through cascaded layers of non-linear convolution and pooling operations. The face representation that emerges  at the top of the network  is a compact vector with an impressive  capacity for robust face recognition.

Although deep networks have made significant progress on the problem of generalized face recognition, the face representation they create  is poorly understood \cite{Sejnowski201907373}. Approaches to dissecting this representation have been aimed at: a.) uncovering the information retained in the descriptor  \cite{parde2017face,hong2016explicit}, b.) probing the robustness of individual unit responses to image variation \cite{parde2017face}, c.) visualizing the receptive fields of {\it individual units} in the network code \cite{qin2018convolutional}, and d.) visualizing the similarity structure of a population of {\it ensemble face representations} for images and identities \cite{hill2019deep}. We consider each in turn.

First, it is now clear that face descriptors from DCNNs retain a surprising amount of information about the original input image \cite{parde2017face}. Specifically, the output representation from DCNNs trained for face identification can be used to predict the viewpoint (yaw and pitch) and media type (still or video image) of the input image with high accuracy \cite{parde2017face,o2018face}. Therefore, deep networks achieve robust identification, not by filtering out image-based information across layers of the network, but by effectively managing it (cf. also \cite{dicarlo2007untangling, hong2016explicit}).

Second, given that DCNN descriptors contain image information, it is possible that the top-layer units separate identity and image information across different units of the face descriptor. Parde et al. (2017) tested this by probing the response properties of the top-layer units in a DCNN trained for face identification to either front-facing or three-quarter-view images of faces \cite{parde2017face}. Individual units did not respond consistently in either a view-specific or view-independent manner.

The third approach is to visualize the response preferences of units in the network \cite{qin2018convolutional} with the goal of translating them  into perceptible images. This is done with techniques such as Activation Maximization \cite{erhan2009visualizing} or deconvolution \cite{zeiler2011adaptive}. This approach is useful for interpreting hidden units at lower layers of DCNNs, where unit activations can be linked to locations within an input image. At higher levels of the network, however, unit responses are not bound to image locations and so
image-based visualization is of limited value.

The fourth approach is to visualize the similarity structure of  {\it ensembles} of DCNN unit activations.  This reveals a highly organized {\it face space}, cf. \cite{valentine1991unified,o2018face}. Specifically, visualization was applied to highly controlled face images of multiple identities that vary in viewpoint and illumination \cite{hill2019deep}. The resulting face space showed that images clustered by identity, identities separated into regions of male and female faces, illumination conditions (ambient vs. spotlight) nested within identity clusters, and viewpoint (frontal to profile) nested within illumination conditions. Therefore, from deep networks trained with in-the-wild images, a highly structured representation of identity and image variation emerges.  

Approaches to examining DCNN codes have focused either on single-unit responses or on the full ensemble of units. Neither provides a complete account of how unit responses interact in a representational space to code semantic information about faces. Here, we examined the juxtaposition of unit and ensemble codes of face identity, gender, and viewpoint in a deep network. At the unit level, we probed the distribution of information about faces and images across individual units in the network's face descriptor. Specifically, we tested identification, gender classification, and viewpoint estimation in variably-sized, randomly sampled, subspaces of top-layer units from a face-identification DCNN. We examined the minimum number of units needed to perform each of these tasks, as well as the predictive power of each individual unit. At the ensemble level, we examined  identity, gender, and viewpoint codes by interpreting them as directions in the representational space created at the top-layer of a DCNN. We did this by performing principal component analysis (PCA) on the face-descriptor vectors and then analyzing the quality of identity, gender, and viewpoint information coded by each PC. The results indicate that identity and image information separate in the ensemble space, but are confounded in individual unit responses. This challenges classical tuning analogies for neural units at high levels of visual processing.

\begin{figure}
	\begin{center}
        \includegraphics[width=3.0in]{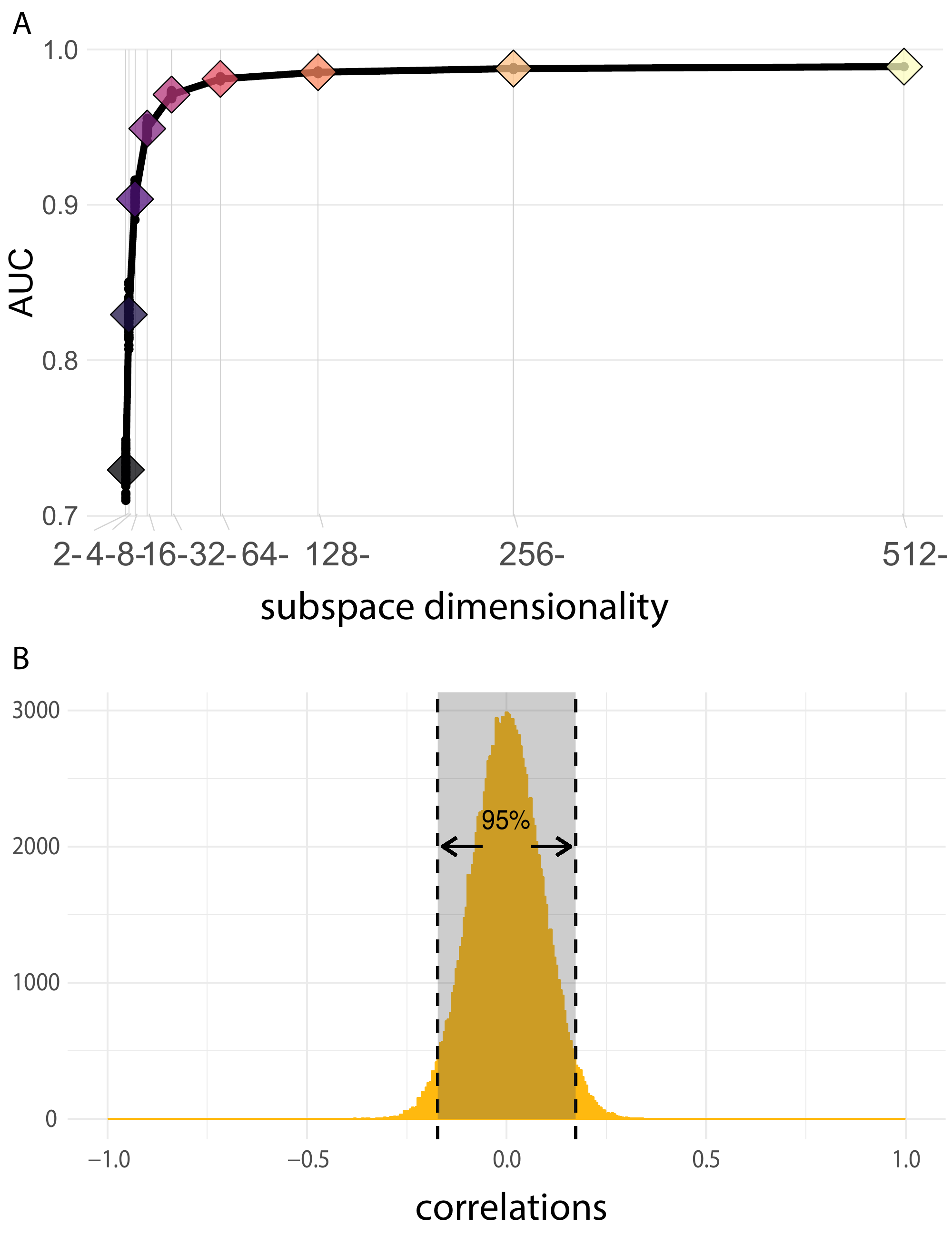}
	    \caption{Identification accuracy is plotted as a function of subspace dimensionality, measured as  area under the ROC curve (AUC) (A). Performance is nearly perfect (AUC $\approx$ 1.0) with the full 512-dimensional descriptor and shows negligible declines until subspace dimensionality reaches 16-units. Performance with as few as two units remains above chance. Correlation histogram for unit responses across images indicates that units capture non-redundant information for identification (B).}
	    
	    \label{fig:identityResults}
	\end{center}
\end{figure}

\section{Results}
\label{sec:headings}

\subsection{Unit-level Face Information}
Face representations were obtained by processing face images ({\it n} = 22,357) of 3,531 identities through a state-of-the-art DCNN trained for face identification. The final face-image representation was defined as the 512-dimensional, penultimate layer of the network. The distribution of identity, gender, and viewpoint was examined across units in randomly sampled subspaces of varying dimensionalities (512, 256, 128, 64, 32, 16, 8, 4, and 2 units). For each dimensionality, 50 random samples were selected.

\subsubsection{Identity}

\paragraph{Identification accuracy is robust in low-dimensional subspaces.} Figure~\ref{fig:identityResults}A shows face-identification accuracy as a function of the number of randomly-selected units sampled from the face representation. Face-identification accuracy is near-perfect in the full-dimensional space. A substantial number of units can be deleted with almost no effect on accuracy for identifying the 3,000+ individuals in the test set. The first substantial drop in performance is seen at 16 units ($\approx3 \%$ of the full dimensionality). Accuracy remains high (AUC = 0.80) with as few as four units, and is well above chance (AUC = 0.72) with only two units. DCNN performance is robust, therefore, with very small numbers of units. Further, performance does not depend on the particular units sampled.

The remarkable stability of identification performance with random selections of very few top-layer units is consistent with two types of codes. First, it is possible that individual units provide diverse cues to identity. Combined, these cues could accumulate to provide a powerful code for identification. If this is the case, we would expect many individual units to show a measurable capacity for separating identities. Moreover, the identity information captured by individual units would be uncorrelated. Alternatively, it may be that many different units capture redundant, but effective, information for face identification. By this account, we expect the response patterns of a subset of units to be highly correlated.

\paragraph{Individual units yield diverse, not redundant, solutions for identity.}Figure ~\ref{fig:identityResults}B shows the distribution of response correlations for all possible pairs of top-level units across all images in the test set. The distribution is centered at zero, with 95\% of correlations falling below an absolute value of 0.17. Therefore, units in the DCNN capture non-redundant identity information.

\paragraph{Highly-distributed identity information across units.}We quantified the identification capacity of individual units in the DCNN. Units with high identification capacity support maximal identity separation while simultaneously minimizing the distance between same-identity images. Therefore, a unit has identity-separation power when its responses vary more between images of different identities than within images of the same identity. We applied analysis of variance (ANOVA) to each unit's responses to all images in the test set. For each ANOVA, identity was the independent variable, image was the random (observation) variable, and the unit responses were the dependent variable. The resulting $F$ ratios provide an index of between-identity variance to within-identity variance. All units had sufficient power to separate identities ({\it p} $<$ .000098, Bonferroni corrected for $\alpha$ = 0.05). 

\begin{figure}
	\begin{center}
		\includegraphics[width=5 in]{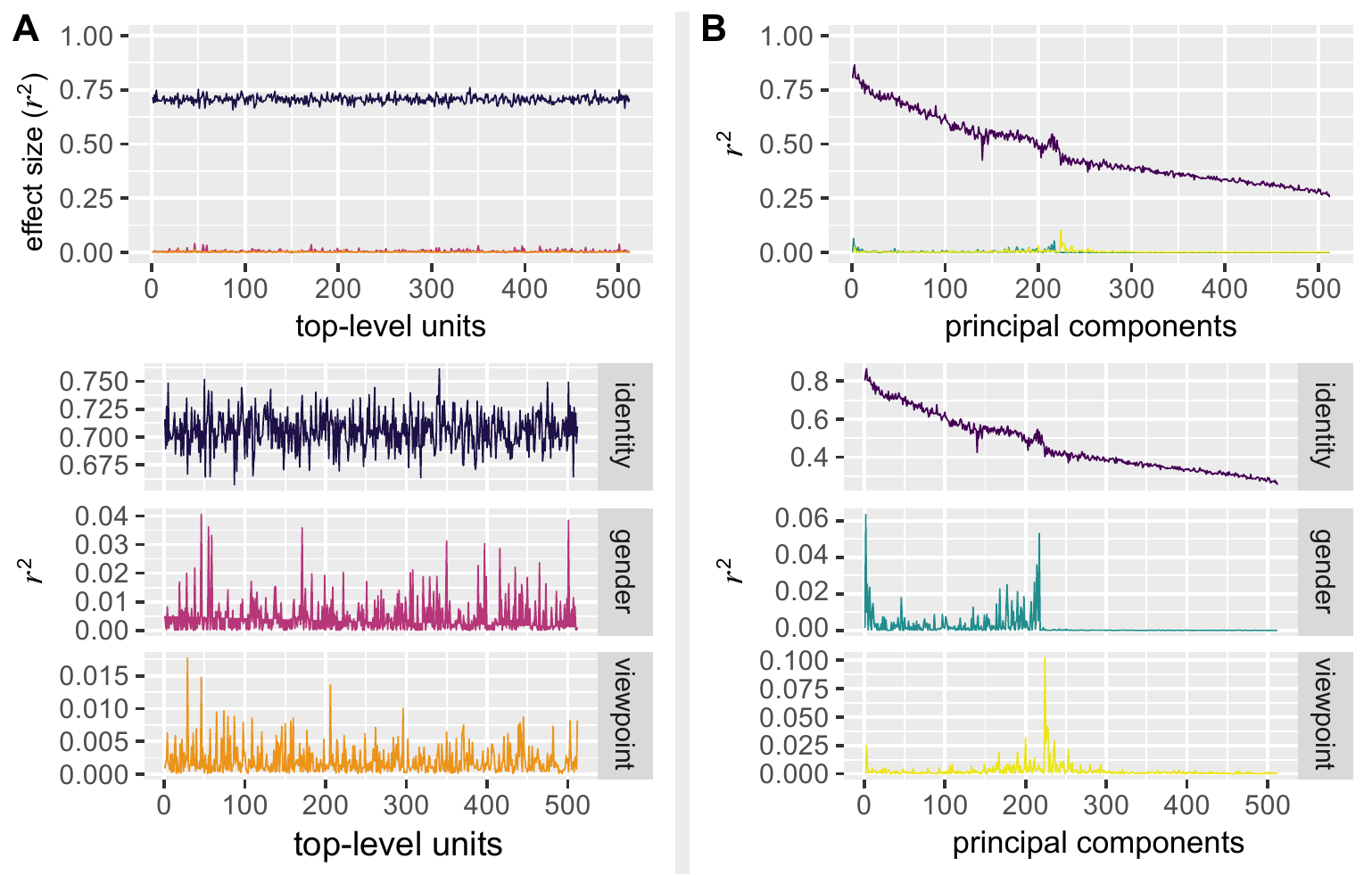}
		\caption{Effect sizes for units (A) and principal components (B) for identity, gender, and viewpoint. For units and principal components, the top panels illustrate the dominance of identity over gender and viewpoint. Lower panels show an approximately uniform distribution of effect sizes for units (A) and differentiated effect sizes for principal components (B) for all three attributes.}
	\label{fig:PCA_all_effect_sizes}
	\end{center}
	\vspace{-0.55cm}
\end{figure}

Next, we calculated the proportion of variance in a unit's response explained by identity variation ($r^2$ effect size). Figure \ref{fig:PCA_all_effect_sizes}A (purple) shows the distribution of effect sizes across units, with an average of $r^2 = 0.691$ (minimum $=$ 0.6573, maximum  $=$ 0.7611). Thus, on average, 69.1\% of the variance in individual unit response is due to variation in identity. All units have a substantial capacity for separating identities.

\subsubsection{Gender}
\begin{figure*}
	\begin{center}
	    \includegraphics[width=3.0in]{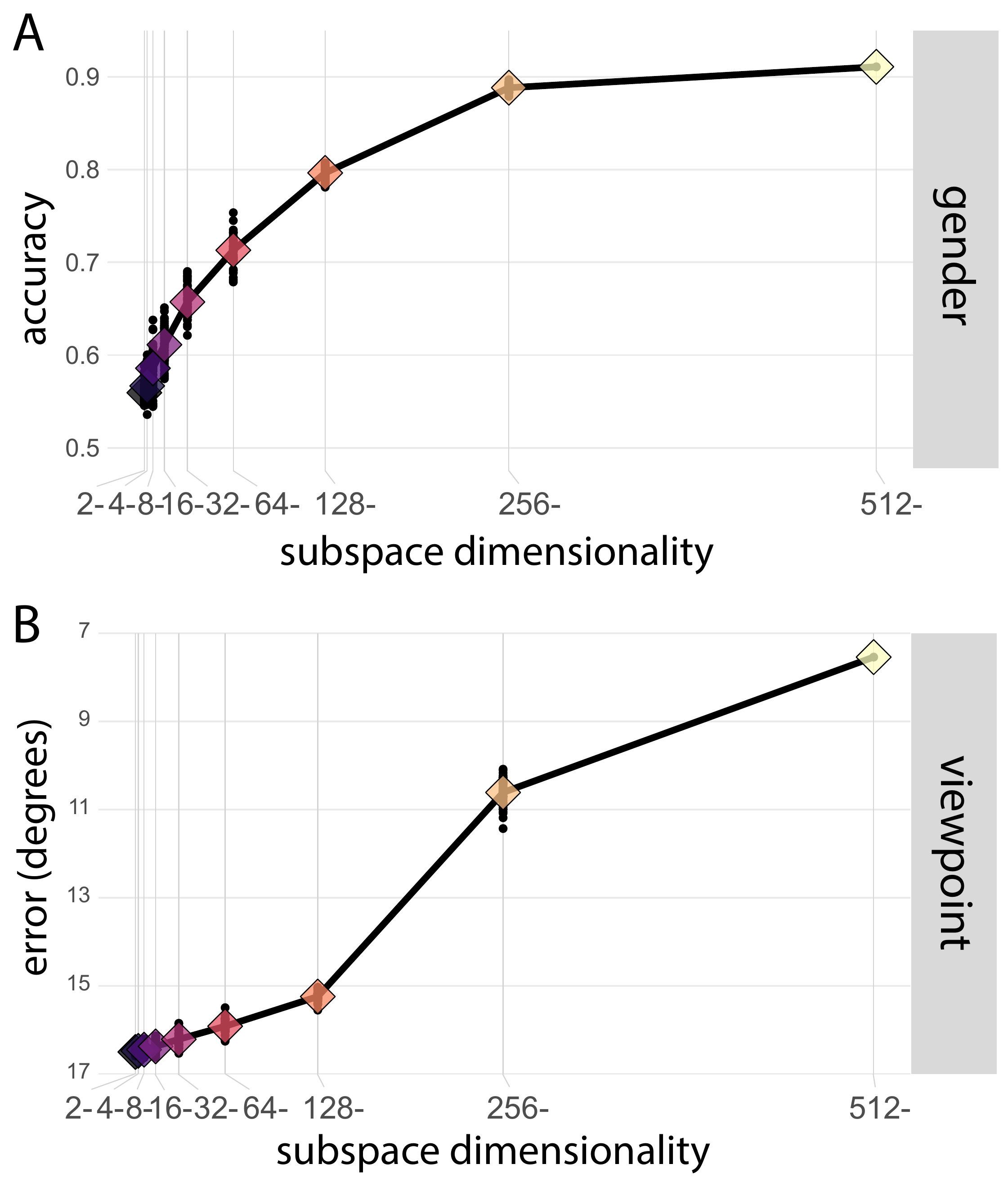}
		\caption{Gender and viewpoint prediction with variable numbers of sampled features. Gender classification declines gradually (A) and viewpoint prediction declines rapidly (B) as sample size decreases. Mean performance for samples ($n = 50$) is shown with a diamond. Different sample sizes appear in different colors.}
		\label{fig:gender_view_pred}
	\end{center}
	\vspace{-0.75cm}
\end{figure*}

Gender-prediction accuracy was measured in the abbreviated subspaces sampled for the face-identification experiments. For each sample, linear discriminant analysis (LDA) was applied to predict the labeled gender (male or female) of all images in the test set from the unit responses. Using all units, gender classification was 91.1$\%$ correct. Classification accuracy declined steadily as the number of units sampled decreased (Figure ~\ref{fig:gender_view_pred}A).

Next, the gender-separating capacity of each individual unit was measured. An ANOVA was performed for each unit, using gender as the independent variable. Overall, 71.5\% of the units were able to separate images according to gender ({\it p} $<$ 0.000098, Bonferroni corrected for $\alpha$ = 0.05). However, gender accounted for only a very small amount of the variance in unit responses. Figure \ref{fig:PCA_all_effect_sizes}A (pink) shows effect size across units (mean $r^2$ = .0045, minimum $\approx$ 0, maximum = 0.041). Notwithstanding the small effect sizes, the finding that 71.5\% of the units' responses differed significantly as a function of gender is meaningful. If individual units did not possess predictive power for gender, approximately 5\% ($\alpha$ level) of units would reach significance.

Consequently, far more units are needed to predict face gender than to predict identity. This is because fewer units have predictive power for gender than for identity, and because the predictive value of these units for gender is weaker.

\subsubsection{Viewpoint}
Viewpoint was predicted using linear regression, cross validated by identity subgroups, and assessed in the same samples tested for identification and gender classification. Prediction error was defined as the difference between predicted and true yaw (in degrees). Figure ~\ref{fig:gender_view_pred}B shows prediction error as a function of the number of randomly-sampled units. Using all units, viewpoint was predicted to within 7.35 degrees. Prediction accuracy was at chance when subspace dimensionality fell to 32 units. Accurate prediction required nearly half of the units.
 
Viewpoint separation capacity for each  unit was assessed with ANOVA, using viewpoint as the independent variable. Effect size measures the proportion of variance explained by viewpoint in each unit's response. Figure \ref{fig:PCA_all_effect_sizes}A (orange) shows small effect sizes for viewpoint  (average $r^2$ = 0.0020, minimum $\approx$ 0, maximum = 0.018). However, overall, 54.7\% of units separated images according to viewpoint ({\it p} $<$ 0.000098, Bonferroni corrected for $\alpha$ = 0.05). Therefore, viewpoint prediction requires far more units than identity or gender prediction.

\subsubsection{Single Unit Summary} Multiple, qualitatively different codes co-exist within the same set of DCNN top-layer units. These codes are differentiated by the number of units needed to perform a task, and by the predictive power of individual units for the task. First, all units provide strong cues to identity that are largely uncorrelated. Therefore, small numbers of randomly chosen units can achieve robust face identification. Second, gender is coded weakly in approximately 72\% of the units. Accurate gender prediction requires a larger number of units, because the set must include gender-predictive units, and these units must be combined to supply sufficient power for classification. Third, even fewer units (about 50\%) code viewpoint---each very weakly. Therefore, a large number of units is needed for accurate viewpoint estimation. 

\subsection{Ensemble Coding of Identity, Gender, and Viewpoint}
How do ensemble face representations encode identity, gender, and viewpoint in the high-dimensional space created by the DCNN? To understand unit-based face-image codes in the context of directions in this  space requires a change in vantage point. A {\it face space representation} \cite{valentine1991unified,o2018face} was generated by applying PCA to the ensemble unit responses. The axes of the space (PCs) are ordered according to the proportion of variance explained by the ensemble face-image descriptors. We re-expressed each face-image descriptor as a vector of PC coordinates. This captures a face-image representation in terms of its relationship to principal directions in the ensemble space.

\begin{figure}
	\begin{center}
		\includegraphics[height=2.6in]{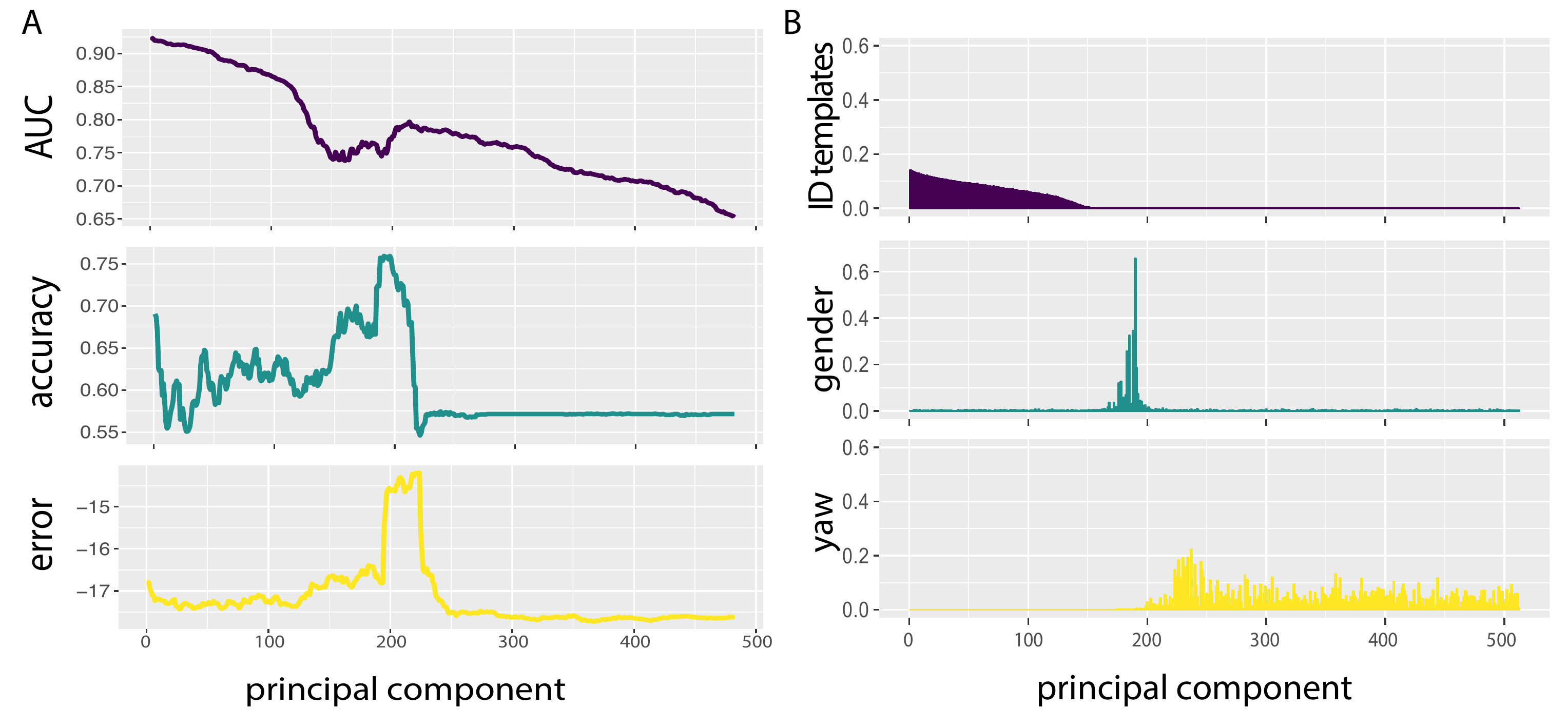}
		\caption{(A) Sliding windows of PCs were then used to predict identity (purple), gender (teal), and yaw (yellow) across the full set of PC subspaces. Identification accuracy is highest when using early PCs. Gender and viewpoint classification accuracy were highest when using subspaces with the highest effect sizes for gender and viewpoint separation, respectively. (B) Similarity between principal component eigenvectors and the directions for identity (purple), gender (teal), and yaw (yellow). The identity direction is the average similarity between identity templates and PCs. The gender direction is the linear discriminant line from an LDA for gender classification. The viewpoint direction is the vector of weights from a linear regression for predicting viewpoint.}
		\label{fig:PCA_analysis}
	\end{center}
	\vspace{-0.75cm}
\end{figure}

For each PC, we measured identity, gender, and viewpoint separation using the ensemble-based image code. Effect sizes were computed for the PC-based codes using ANOVA, as was done for the unit-based codes. These appear in Figure~\ref{fig:PCA_all_effect_sizes}B. Identity (purple) dominates gender (teal) and viewpoint (yellow) information, consistent with the unit-based code (cf. Fig. \ref{fig:PCA_all_effect_sizes}A). In  contrast to the unit-based codes, effect sizes for individual PCs are strongly differentiated by face attribute. Effect sizes for identity are highest in PCs that explain the most variance in the ensemble space. Gender information peaks in two ranges of PCs ($\approx$2--10 and $\approx$164--202). For viewpoint, effect sizes peak between PCs $\approx$220 and 230. Therefore, face-image attributes are filtered into multiple subspaces and are ordered roughly according to explained variance.

Next, we show that these subspaces differ in their functional capacity to classify identity, gender, and viewpoint. Moreover, these subspaces align with directions in the representational space diagnostic of face attributes.

\paragraph{Face Attributes Predicted from Ensemble Codes}
To test the ability of different subspaces to separate attributes, we predicted identity, gender, and viewpoint from different ranges of PCs. Starting with PCs 1 to 30, we used sliding windows of 30 PCs (1--30, 2--31, 3--32, etc.) to predict each face-image attribute. Figure~\ref{fig:PCA_analysis}A shows that the accuracy of the predictions for the three attributes differs with the PC range. Identification accuracy is best in the subspaces that explain the most variance. Gender-classification accuracy is highest when using ranges of PCs that encompass the highest effect sizes for gender separation. Similarly, viewpoint prediction is most accurate with ranges of PCs that encompass the highest effect sizes for viewpoint separation.

\paragraph{Face Attributes Align with Directions in the Space}
At a more general level, it is possible to compare the PC directions in the space to the directions diagnostic of identity, gender, and viewpoint. We compared PCs to: a.) the directions of identity codes, b.) the direction in the space that maximally separated faces by gender (gender direction), and c.) the direction that best supported viewpoint prediction (viewpoint direction). Identity codes were created by averaging the face descriptors for all images of an identity. The gender direction was the linear discriminant line from the LDA used for gender classification. The viewpoint direction was the vector of regression coefficients for viewpoint prediction.

Figure~\ref{fig:PCA_analysis}B (purple) shows the average of the absolute value of cosine similarities between each PC and all identity codes. Figure~\ref{fig:PCA_analysis}B (teal) shows the similarity between each PC and the gender direction, and Figure~\ref{fig:PCA_analysis}B (yellow) shows the similarity between each PC and the viewpoint direction. These plots reveal that identity information is distributed primarily across the first $\approx$150 PCs, gender information is distributed primarily across PCs ranked between $\approx$150--200, and viewpoint information is distributed primarily across PCs ranked greater than $\approx$200. 

Consistent with the effect sizes computed for each PC, as well as the attribute predictions, this result shows that identity, gender, and viewpoint are filtered roughly into subspaces ordered according to explained variance in the DCNN-generated ensemble space. This filtering reflects a prioritization of identity over gender, and of gender over viewpoint.

\begin{figure}
	\begin{center}
		\includegraphics[width=3.0in]{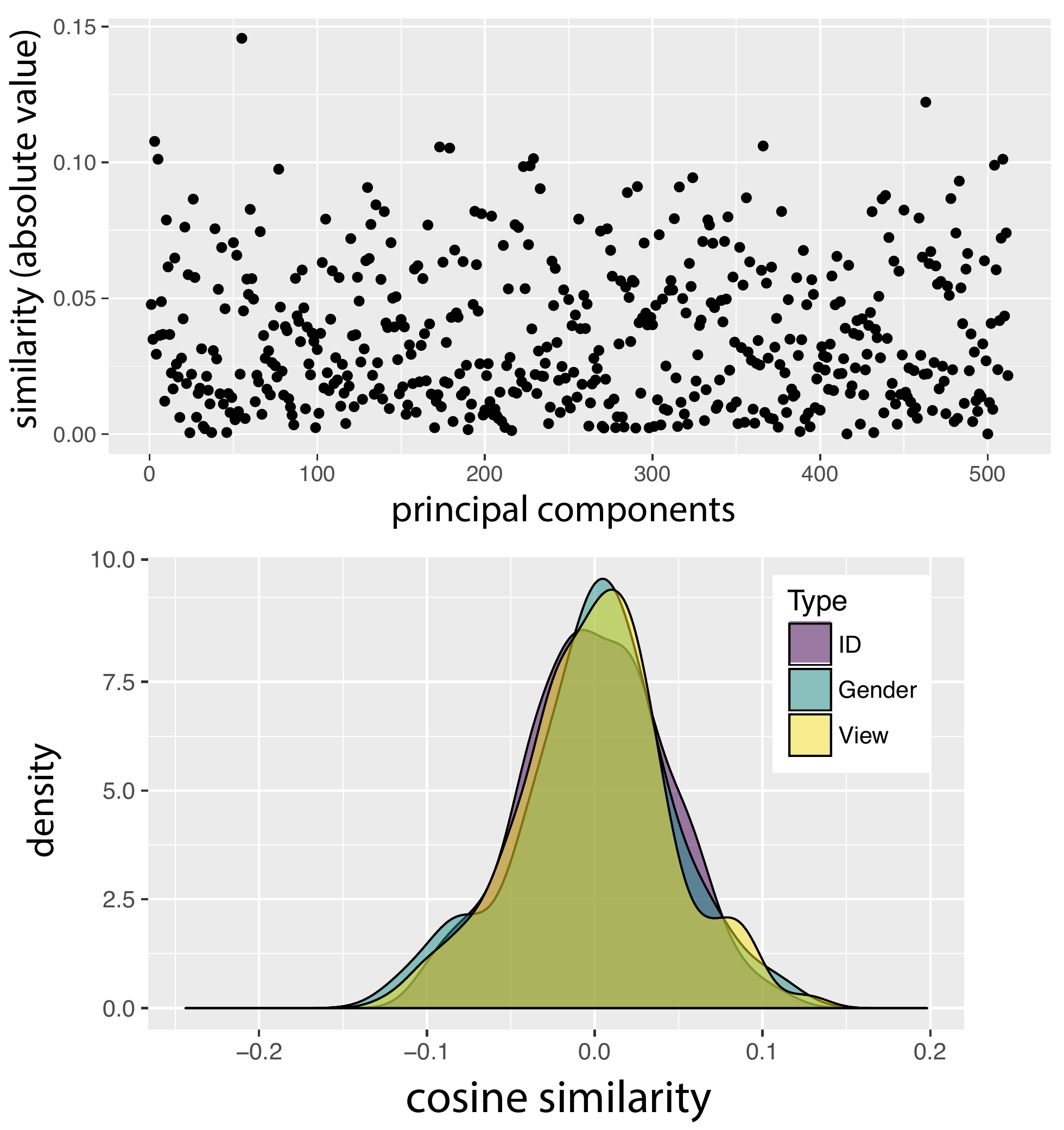}
		\caption{For a single example unit, absolute value of similarities between unit direction and each principal component shows confounding of unit response with identity, gender, and viewpoint (top). Density plot of similarities between the example unit and principal components associated with identity (purple), gender (blue), and viewpoint (yellow) (bottom). The distributions overlap almost completely, indicating that each type of information contributes to the unit's activation. This finding was consistent across all unit basis vectors.}
		\label{fig:unit_PCA_basis}
	\end{center}
\end{figure}

\subsection{Juxtaposed Unit and Ensemble Codes} 
PCs capture directions that can be interpreted in terms of identity, gender, and viewpoint. How do these directions relate to the basis vectors that define the DCNN units? This will tell us whether individual units ``respond preferentially'' to directions that can be interpreted in terms of identity, gender, or viewpoint. 

We calculated the cosine similarity between the PC directions and the unit directions (unit 1: $[1,0,0 ... 0_{512}]$, unit 2: $[0,1,0 ... 0_{512}]$, etc.). If a unit responds preferentially to viewpoint, gender, or identity, it will align closely with PCs related to a specific attribute. Confounding of  semantically relevant information (identity, gender, viewpoint) in a unit's response will yield a uniform distribution of similarities across the PCs.

\paragraph{Unit responses confound identity, gender, and viewpoint.}Figure~\ref{fig:unit_PCA_basis} (top) shows a uniform distribution of similarities across PCs for a single unit. We found this for all of the 512 units (see Supplemental Information). Figure~\ref{fig:unit_PCA_basis} (bottom) shows the histogram of these similarities, separated by attribute. Identity, gender, and viewpoint information, which are separated in the high dimensional space, are confounded in unit responses. This undermines a classic tuning analogy for units. In isolation, individual units cannot be interpreted in terms of a specific identity, gender, or viewpoint.

\section{Discussion}
Historically, neural codes have been characterized as {\it either} sparse or distributed. Sparse codes signal a stimulus via the activity of a small number of highly-predictive units. Distributed codes signal a stimulus via the combined activity of many weakly-predictive units. The DCNN's identity code encompasses fundamental mechanisms of both sparse (highly predictive single units) and distributed (powerful combinations of units) codes. This unusual combination of characteristics accounts for the DCNN's remarkable resilience to unit deletion. Superimposed on the identity representation are standard distributed codes for gender and viewpoint, and likely other subject and image variables. For these codes, ensembles, not individual units, make accurate attribute predictions.

The results reveal three distinct attribute codes (identity, gender, view) in one set of units. These codes vary in the extent to which they distribute information across units. Because multiple attribute codes share the same units, the labels ``sparse'' or ``distributed'' must specify a particular attribute. In deep layers of DCNNs, where units respond to complex combinations of low-level visual features, these shared codes may be common. If these codes exist in the primate visual system, it would likely be at  higher-levels of the visual processing hierarchy. In low-level visual areas (e.g., V1), neural receptive fields reference to locations in the retinotopic image, and are more likely to act as single-attribute ``feature detector'' codes.

Much of what appears complex in individual units is clear in the ensemble space. PCs separate attributes in the DCNN representation according to explained variance. This reflects network prioritization (identity $>$ gender $>$ viewpoint). PCs comprise a ``special'', interpretable, rotation of the unit axes, because the face attributes are not represented equally in the face descriptors. The juxtaposition of unit and ensemble characteristics indicates that information coded by a deep network is in the representational space, not in any given projection of the space onto a particular set of axes, cf. \cite{szegedy2013intriguing}. 

How then are we to understand the units? The DCNN is optimized to separate identity, not to maximize the interpretability of the information that achieves identity separation. From a computational perspective, any orthogonal basis is as good as any other. Given the high dimensionality of the system and the stochastic nature of training, the likelihood of a DCNN converging on a semantically interpretable basis set is exceedingly low. Units serve the sole purpose of providing a set of basis axes that support maximal separation of identities in the space. There should be no expectation that the response of individual units be ``tuned'' to semantic features. In isolation, units provide little or no information about the visual code that operates in the high-dimensional space. Instead, units must be interpreted in an appropriate population-based computational framework \cite{chang2017code,hong2016explicit,yamins2014performance}.

How does this affect the way we interpret neural data? The literature is replete with reports of preferentially-tuned neurons in face-selective cortex. Electrophysiological recordings differentiate face patches, based on the tuning characteristics of neurons (e.g., PL: eyes, eye region, face outlines, \cite{issa2012precedence}; ML: iris size, inter-eye distance, face shape, and face views \cite{freiwald2010functional}; AM: view-invariant identity \cite{chang2017code,freiwald2010functional}). The problem with interpreting the responses of single neurons is evident when we consider what a neurophysiologist would conclude by 
recording from top-layer units in the network we analyzed. 

First, most of these units would appear to be ``identity-tuned'', preferring some identities (high activation) over others (low activation). However, our data show that each unit exhibits substantial identity-separation capacity (cf. effect sizes). Effect sizes consider the full range of responses, instead of making only a ``high'' versus ``low'' response comparison. The neural-tuning analogy obscures the possibility that individual units can contribute to identity coding with a relatively low-magnitude response. This response, in the context of the responses of other units, {\it is} information in a distributed code. A neurophysiologist would find ``identity-tuned units'' here (in what is, essentially, a distributed code), {\it only} because identity modulates the individual unit responses so saliently.  These are not identity-tuned units, they are identity-separator units. Moreover, what separates identity in these units is not likely to be interpretable. This is due to the uncertain relationship between meaningful directions in the representational space and the arbitrary directions of the unit axes.

Second, no units would appear to be tuned to gender or viewpoint, because these attributes modulate the response of a unit only weakly in comparison to identity. As a consequence, the distributed codes that specify gender and viewpoint would be hidden, despite the fact that the ensemble of units contains enough information for accurate classification of both attributes. The hidden modulation of unit responses by viewpoint would, from a neural tuning perspective, imply that the units signal identity in a viewpoint-invariant way. This is, in fact, correct, but provides a misleading characterization of the importance of these units for encoding viewpoint.

Neurophysiological investigations of visual codes typically rely on neural-tuning data from single units in conjunction with population decoding methods. However, if DCNN-like codes exist in primate visual cortex, over-emphasis on neural-tuning functions in high-level areas may be counter-productive. Rather than characterizing neural units by the features or stimuli to which they respond \cite{bashivan2019neural}, we should instead consider units as organizational axes in a representational space \cite{szegedy2013intriguing}. The importance of a unit lies in its utility for separating items within a class, not in interpreting the attributes for which it has a high or low activation. This requires a shift in perspective from the principles of sparse and distributed coding that, although helpful for understanding early visual processing, might not be appropriate in high-level vision.

\section{Methods}
\label{sec:methods}

\subsection{Network}
All reported data are from a 101-layered face-identification DCNN \cite{ranjan2017all}. This network performs with high accuracy across changes in viewpoint, illumination, and expression (cf. performance on IARPA Janus Benchmark-C [IJB-C] \cite{Maze2018IARPAJB}). Specifically, the network is based on the ResNet-101 \cite{wen2016discriminative} architecture. It was trained with the Universe dataset \cite{bansal2017s,ranjan2018crystal}, comprised of three smaller datasets (UMDFaces \cite{bansal2017umdfaces}, UMDVideos \cite{bansal2017s}, and MS-Celeb-1M \cite{guo2016ms}). The dataset includes 5,714,444 images of 58,020 identities. The network employs Crystal Loss (L2 Softmax) for training \cite{ranjan2018crystal}. Crystal Loss scale factor $\alpha$ was set to 50. ResNet-101 employs skip connections to retain the strength of the error signal over its 101-layer architecture. Once the training is complete, the final layer of the network is removed and the penultimate layer (512 units) is used as the identity descriptor. This penultimate layer is considered the ``top layer'' face-image representation.
	
\subsection{Test Set}
The test set was comprised of images from the IJB-C dataset, which contains 3,531 subjects portrayed in 31,334 still images (10,040 non-face images) and frames from 11,779 videos. For the present experiments, we used all still images in which our network could detect at least one face, and for which viewpoint information was available. In total, we selected 22,248 (9,592 female; 12,656 male) faces of 3,531 (1,503 female; 2,028 male) subjects. Note that several images contain multiple detectable faces.

\subsection{Identity AUC Calculation}
For face-identification, images from the test set were assigned randomly to Group A or B. AUCs were computed by comparing every image in set A (5,562 images of 3,056 identities) to every image in set B (5,562 images of 3,053 identities). In total, each AUC was computed from 30,935,844 comparisons.

\subsection{Classification}

\subsubsection{Gender}
Linear discriminant analysis (LDA) was used to classify face gender for each image in the dataset. For each subset of features, a LDA was trained using all images of 3,231 identities and tested on the remaining 300 identities. This process was repeated, holding out a different set of 300 identities each time until all images were classified. Gender labels for each identity were verified by human raters. The final output values were categorical gender labels that could be compared directly to the ground-truth data.

\subsubsection{Viewpoint}
Viewpoint was predicted using linear regression. Linear regression models were computed using the Moore-Penrose pseudo-inverse. For each subset of features, a regression model was trained using all images of 3,231 identities and tested on the remaining 300 identities. This process was repeated, each time holding out a different set of 300 identities, until viewpoint predictions had been assigned to each image. Ground truth for viewpoint was produced by the Hyperface system \cite{chen2016unconstrained} and was defined as the deviation from a frontal pose, measured in degrees yaw (i.e., 0 = frontal, 90 = right profile, -90 = left profile). Output predictions were continuous values corresponding to the predicted viewpoint in degrees yaw.
	
\subsubsection{Permutations}
Permutation tests were used to evaluate the statistical significance of the viewpoint and gender predictions. A null distribution was generated from the original data by randomly permuting values within each unit. Predictions made from the resulting permutations ($n=1,000$) were compared to the true values from each classification test. All permutation tests were significant at $p < 0.001$, with no overlap between test value and null distribution.

\subsection{Analysis of Variance}
For each face-image attribute (identity, gender, viewpoint), an ANOVA was computed for each of the 512 units in the top-level DCNN output, as well as for all 512 PCs. For each ANOVA, the independent variable was either the vector of a unit's responses or the vector of a PC's factor scores. Face images were used as the random variable. Identity, gender, and viewpoint were used in separate analyses as the dependent variable. For viewpoint, the absolute value was binned into the following five categories. Frontal: [0$^{\circ}$, 18$^{\circ}$]; near-frontal: (18$^{\circ}$, 36$^{\circ}$]; half profile (36$^{\circ}$, 54$^{\circ}$]; near profile (54$^{\circ}$, 72$^{\circ}$]; profile (72$^{\circ}$, 150$^{\circ}$]. To account for unequal group sizes, pooled sums-of-squares were used as the error term.

\section*{ACKNOWLEDGMENT}
Funding provided by National Eye Institute Grant R01EY029692-01 and by the Intelligence Advanced Research Projects Activity (IARPA) to AOT. This research is based upon work supported by the Office of the Director of National Intelligence (ODNI), Intelligence Advanced Research Projects Activity (IARPA), via IARPA R\&D Contract No. 2014-14071600012 and 2019-022600002. The views and conclusions contained herein are those of the authors and should not be interpreted as necessarily representing the official policies or endorsements, either expressed or implied, of the ODNI, IARPA, or the U.S. Government. The U.S. Government is authorized to reproduce and distribute reprints for Governmental purposes notwithstanding any copyright annotation thereon.

\bibliographystyle{unsrt}  
\bibliography{references}

\end{document}